\documentclass[a4paper, twoside, 12pt]{report}

\usepackage[english]{babel}
\usepackage[utf8x]{inputenc}
\usepackage[T1]{fontenc}

\usepackage[a4paper,top=3cm,bottom=2cm,left=3cm,right=3cm,marginparwidth=1.75cm]{geometry}

\usepackage{amsmath}
\usepackage{float}
\usepackage{commath}
\usepackage{amssymb}
\usepackage{graphicx}
\usepackage[table]{xcolor}
\usepackage{array}
\graphicspath{ {figures/} }
\usepackage[colorinlistoftodos]{todonotes}
\usepackage[colorlinks=true, allcolors=blue]{hyperref}
\usepackage[toc,page]{appendix}

\title{Machine Learning for Health: Personalized Models for Forecasting of Alzheimer Disease Progression}
\author{Aritra Banerjee\\CID 01539378}

\begin{document}
\begin{titlepage}

\newcommand{\HRule}{\rule{\linewidth}{0.5mm}} 


\includegraphics[width=8cm]{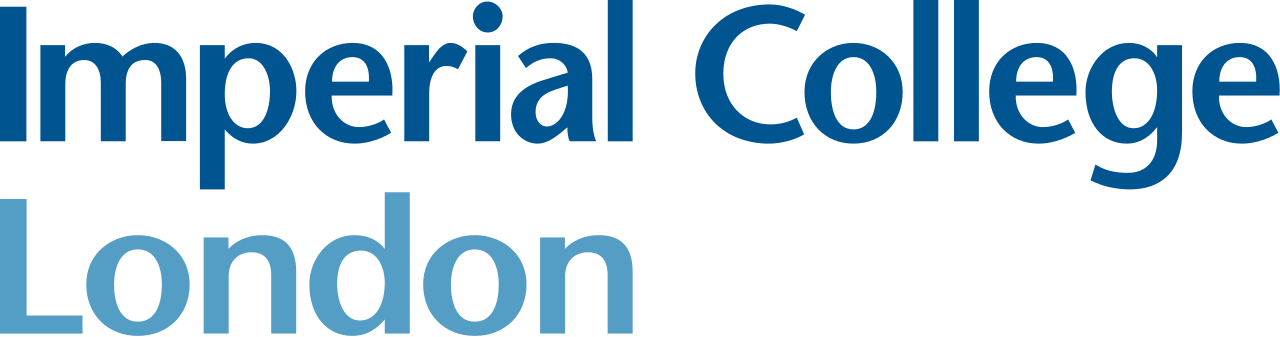}\\[1cm] 
 

\center 


\textsc{\LARGE Msc Individual Project(Machine Learning)}\\[1.5cm]
\textsc{\Large Imperial College London}\\[0.5cm] 
\textsc{\large Department of Computing}\\[0.5cm] 

\makeatletter
\HRule \\[0.4cm]
{ \huge \bfseries \@title}\\[0.4cm] 
\HRule \\[1.5cm]
 

\begin{minipage}{0.4\textwidth}
\begin{flushleft} \large
\emph{Author:}\\
\@author 
\end{flushleft}
\end{minipage}
~
\begin{minipage}{0.4\textwidth}
\begin{flushright} \large
\emph{Imperial Supervisor:} \\
Prof. Daniel Rueckert \\[1.2em] 
\emph{MIT Supervisor:} \\
Dr. Ognjen Rudovic 
\end{flushright}
\end{minipage}\\[2cm]
\makeatother



{\large \today}\\[2cm] 

\vfill 

\end{titlepage}

\begin{abstract}
In this thesis, the aim is to work on optimizing the modern machine learning models for personalized forecasting of Alzheimer Disease (AD) Progression from clinical trial data. The data comes from the TADPOLE challenge, which is one of the largest publicly available datasets for AD research (ADNI dataset). The goal of the project is to develop machine learning models that can be used to perform personalized forecasts of the participants’ cognitive changes (e.g., ADAS-Cog13 scores) over the time period of 6,12, 18 and 24 months in the future and the change in Clinical Status (CS) i.e., whether a person will convert to AD within 2 years or not. This is important for informing current clinical trials and better design of future clinical trials for AD. We will work with personalized Gaussian processes as machine learning models to predict ADAS-Cog13 score and Cox model along with a classifier to predict the conversion in a patient within 2 years.\\
This project is done with the collaboration with researchers from the MIT Media Lab.
\end{abstract}
\renewcommand{\abstractname}{Acknowledgements}
\begin{abstract}
I would like to thank everyone who have provided particularly useful assistance, technical or otherwise, during my MSc Individual Project (Specialism). I would like to thank my supervisors Dr. Daniel Rueckert and Dr. Ognjen Rudovic specifically for being outstanding mentors and guides throughout the entire project. They guided me through every step and ensured that I work on this thesis according to my full potential within the given deadline.
\end{abstract}
\tableofcontents
\thispagestyle{empty}
\listoffigures 
\listoftables
\newpage
\chapter{Introduction}
\section{General Idea}
Alzheimer's disease is a dangerous neural psychological disease which affects around 30 million people worldwide in the year 2015. It causes the state of dementia which affects the person suffering the disease as well the people around them. The person affected tend to forget the recent events. It usually affects people who are 65 years and older. Approximately $6\%$ of people over 65 years suffer from Alzheimer's\cite{burns2009alzheimer}. Sadly after spending billions of pounds, thousands of clinical trials, only less than 1$\%$ of the clinical trials have successfully gone into the regulatory approval stage and none of them have found a way to prove the disease-modifying effect\cite{marinescu2018tadpole,cummings2006challenges}. \\
In this project, we are going to use personalized machine learning techniques for forecasting of AD Progression. For that the basic concept is taken from one of the papers published jointly by Imperial and MIT researchers\cite{peterson2017personalized}. This paper is the building block on which we will mainly emphasize in this project.\\
In this paper AD forecasting is done on mainly the three metrics:
\begin{itemize}
     \item MMSE\cite{folstein1975mini}: Mini-Mental State Examination.
     \item CDRSB\cite{hughes1982new}:The Washington University Clinical Dementia Rating Sum of Boxes score.
     \item ADAS-Cog13\cite{rosen1984new}: AD Assessment Scale-Cognitive Subset.
\end{itemize}
CS (Clinical Status)\cite{marinescu2018tadpole,weiner2017recent} is the actual conversion metric used to decide whether a person is actually converted as an Alzheimer's patient or not. It can be either Cognitively Normal(CN); Mild Cognitive Impairment(MCI) or Early Alzheimer Disease(AD). From a doctor's point of view these values should always be known for him or her to decide whether a patient has Alzheimer's or not. But for many cases the patient does not have enough recorded data available. That is the main purpose of the our project. To provide the doctor with an idea about the patient's conversion without having an actual ADAS-Cog13 score or not enough CS values.
\begin{figure}[H]
  \centering
  \includegraphics[width=0.75\textwidth]{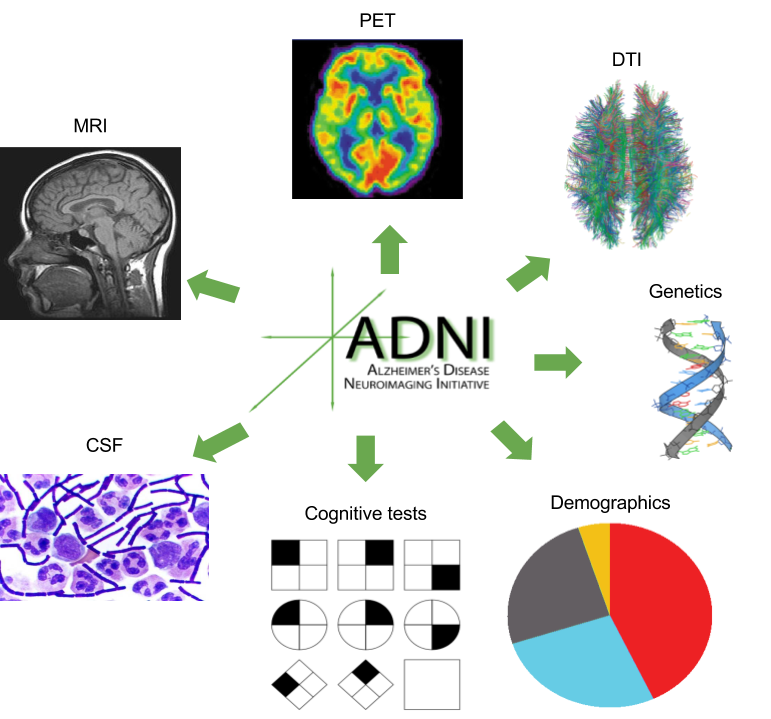}
  \caption{Overview of ADNI Biomarkers\cite{marinescu2018tadpole}.Source of Individual pictures: \href{https://commons.wikimedia.org/wiki/Main_Page}{Wikipedia Commons}}
  \label{fig:adni}
\end{figure}
The multi-modal feature set consists of demographics, cognitive tests, CSF(Cerebrospinal Fluid), MRI(Magnetic Resonance Imaging), PET(Positron Emission Tomography), DTI(Diffusion tensor imaging), Genetics. The ADNI study basically aims to analyze the bio-markers from the different feature sets in order to predict or forecast AD progression. The PET data is removed from the dataset due to sparseness. The cognitive tests consists of mainly the ADAS-Cog13, ADAS-Cog11, CDRSB, MMSE etc. among which the ADAS-Cog13 is the forecast feature that we are trying to work with. Also we have an "Others" feature set which consists of the Clinical Statuses (i.e., the CN, MCI and AD stages). The different features of ADNI dataset can be combined to predict the changes in a person from Cognitively Normal (CN) to Mild Cognitive Impairment (MCI) to stages of early AD which is ultimately the goal of the project. The forecasted ADAS-Cog13 score and CS actually finally helps predict whether a person will convert or not.
\section{Main Focus and Challenges}
For our project we are planning to use the CS and the predicted ADAS-Cog13 features of the dataset to predict whether $(i)$ a person will convert to AD or not and $(ii)$ if the person does convert then how much time it will take for the patient to convert.\\
We can assume a two year sliding window of 4 parts with a gap of 6 months between each visit. So we have four sliding windows. For each window we will calculate the conversion and assign a label to each conversion. If there is no conversion we will not assign any label. Finally for each patient we will see after the 4 sliding windows whether the patient converts to AD or not. If the patient does convert to AD we will also check at what sliding window the patient does convert to AD. Currently we have approximately 1737 patients with a maximum of 21 visits per patient, i.e 10 years. Although for many patients the number of visits are way less and also they do not have a $24^{th}$ month visit at all. This is also a challenge of the project to ensure we have 6,12,18,24 months data for all the patients.\\
It is also imperative to understand that since we are dealing with a lot of patients here sometimes taking all the patients at a time may not be a feasible idea to express its features. Sometimes it is better to take around 500 patients at a time to check a pattern of AD progression and then apply it to all the 1737 patients. Around 1700 patients with around 15 visit approx per person can result in a total of 75k visits which can be very hard to visualize in a system. We will try to calculate average values for different changes in CS values or the ADAS-Cog13 score for around 500 patients (say) and try to generalize if possible. If there is no set pattern we cannot generalize and we may have to calculate for all the patients and all the visits.\\
All patients absolutely have their $0^{th}$ visit registered in the dataset so we will assume that as the baseline to proceed with. If the patient has CS as (CN=1 or MCI=2) we will start the calculations for the possible change of that patient into AD. If the patient has CS as (AD=3) then that patient has already been diagnosed with AD which is currently not within the scope of the project.Sometimes it has been seen that the patient can change back to CN from AD and then again back to AD from CN. Such cases need to be dealt with separately which is currently not being dealt with although we can try to find a feasible solution later on as we advance in the project.
\section{Problem Definition}
As stated above in the introduction section the broad idea is whether a person will convert to AD or not within the next 24 months and if the patient does then how much time it takes to convert that person to AD. We aim to use the non-parametric models like Gaussian processes to solve this problem. \\
In a supervised setting we will use multi-modal features of around 1737 patients as input. The outcome ADAS-Cog13 scores for each patient simultaneously for 6,12,18 and 24 months is based on the multi-modal features stated above in the introduction section. This predicted ADAS-Cog13 score is then used along with the CS of each patient to confirm whether the patient will convert within 2 years or not.
In the dataset we have some patients whose data are missing for a particular visit. For example, we need the data set of each patient to have 6,12,18,24 months of his or her visit being recorded as our ultimate aim is to forecast the AD for these 4 different visits. For those patients that the data is missing currently we will fill them with their past visit data. Missing data is usually represented in our dataset as -999999.\\
First we will perform population level GP\cite{rasmussen2006cki} to train the data. We formulate the forecasting AD problem as a regression setting and solve it using auto-regressive GPs\cite{candela2003propagation}. Next the the population level GP\cite{rasmussen2006cki} is improved by a domain-adaptive GPs\cite{eleftheriadis2017gaussian,liu2015bayesian} approach to personalize the population GP for target patients. This approach we will be using is named personalized GP or pGP\cite{peterson2017personalized}. For each of the first visit of each patient the population model is used instead of the personalized model as it is the source visit. And all the later visits are which are based on the pGP\cite{peterson2017personalized} are based on personalizing the previous visits. From this GP model we will predict the ADAS-Cog13 scores for 6,12,18 and 24 months at the same time. \\
Then, we will use these predicted ADAS-Cog13 scores and change in CS of each patient to predict whether the patient will actually convert within two years or not. This is done by first using the Cox model\cite{cox1972regression} to predict the probabilities of 6,12,18 and 24 months conversion chance. Then using these probabilities and the ground truth of CS conversion we classify whether a patient will actually convert or not.\\
\pagebreak
\section{Flowchart of the entire project}
The below flowchart depicts the entire pipeline of the thesis undertaken:
\begin{figure}[H]
  \centering
  \includegraphics[width=0.25\textwidth]{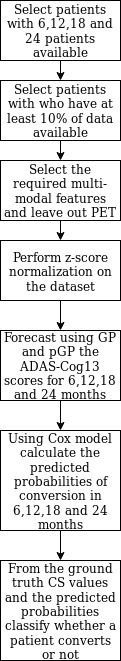}
  \caption{Overview of the entire project pipeline}
  \label{fig:flow}
\end{figure}
\chapter{Background}
The main background idea of the project was based on Rudovic et al. paper \cite{utsumil2018personalized} which extended the approach of Peterson's paper\cite{peterson2017personalized} to predict ADAS-Cog13 score for each patient for a period of 6,12,18 and 24 months. This paper's idea\cite{utsumil2018personalized} was fundamental in predicting the 4 different time frame ADAS-Cog13 scores of the future which in turn helped us predict the chance of a patient converting within 2 years using Cox model\cite{cox1972regression}.\\
Most of the approaches which already exist build on developing models based on their Clinical Status of each patient like in \cite{schmidt2016learning} or \cite{guerrero2016instantiated}. But this paper\cite{peterson2017personalized} initially extends the population level GP into domain adaptive GPs(i.e, the personalized GPs). Then Rudovic's paper \cite{utsumil2018personalized} does this for 6,12,18 and 24 months of ADAS-Cog13 forecast instead of just 6 months again using the pGP and tGP(target GP) approach.
\section{Gaussian Process(GP) Regression Models: The Idea}
Gaussian process is basically a random process which has a collection of random variables such that every finite linear combination of them is a multivariate normal distribution.\\
We have a regression problem to solve using Gaussian processes. The basic idea of Gaussian usage in Regression problems was first stated in the paper\cite{williams1996gaussian} which introduces a Gaussian prior in the Bayesian approach\cite{neal1993bayesian}.
In a Bayesian linear regression setting\cite{mml} let us consider the following model:
\begin{equation}
    \begin{split}
        y = \phi^{T}(x)\theta + \epsilon \\
        \epsilon \sim \mathcal{N}(0,\sigma^{2})\\
        \theta \sim \mathcal{N}(m_{0},S_{0})
    \end{split}
\label{eq:1}
\end{equation}
Here, the Gaussian prior is placed on the parameter $\theta$ as $p(\theta) = \mathcal{N}(m_{0},S_{0})$.\\
We can also represent equation \eqref{eq:1} as a graphical model as below:
\begin{figure}[H]
  \centering
  \includegraphics[width=0.55\textwidth]{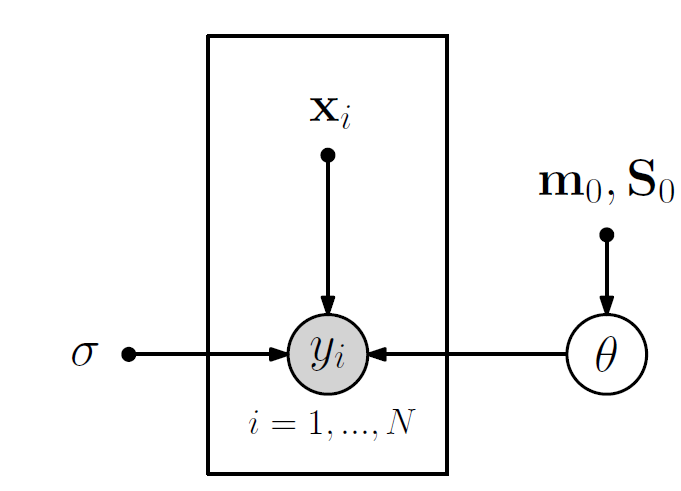}
  \caption{Overview of a Basic Bayesian Regression model}
  \label{fig:bayesian}
\end{figure}
Here, $x_{i} \sim \mathbb{R}^{D}$ which represents the input training set and $y_{i} \sim \mathbb{R}$ are the corresponding set of observations.\\
Now let us calculate the parameter posterior for this model using Bayes' Theorem\cite{d1995multidimensional},
\begin{equation}
    p(\theta|X,y) = \frac{p(y|X,\theta)p(\theta)}{p(y|X)}
    \label{post}
\end{equation}
The $p(y|X,\theta)$ is called the likelihood and $p(\theta)$ is the parameter prior as mentioned above.\\
Also,
\begin{equation}
    p(y|X) = \int p(y|X,\theta)p(\theta)
\end{equation}
is called the marginal likelihood or evidence. This value is independent of the value of the parameter $\theta$.
Now, using the model in equation \eqref{eq:1} we can compute the posterior in equation \eqref{post} in a closed form as:
\begin{equation}
    p(\theta|X,y) = \mathcal{N}(\theta|m_{N},S_{N})
\end{equation}
\begin{equation}
    S_{N} = (S_{0}^{-1} + \sigma^{-2}\Phi^{T}\Phi)^{-1}
\end{equation}
\begin{equation}
    m_{N} = S_{N}(S_{0}^{-1}m_{0} + \sigma^{-2}\Phi^{T}y)^{-1}
\end{equation}
Here, the \textit{N} represents the entire training set.\\
\subsection{GP Posterior Predictions}
Usually a Gaussian process is defined by the mean function \textit{m} and the kernel or co-variance function \textit{k}.
The co-variance or the kernel function should always be positive semi-definite and symmetric.
We can define prediction at any point $x_{*}$ using a GP posterior as,
\begin{equation}
    p(f(x_{*})|X,y,x_{*}) = \mathcal{N}(m_{post}(x_{*}),k_{post}(x_{*},x_{*}))
\end{equation}
where,
\begin{equation}
    m_{post}(x_{*}) = m(x_{*}) + k(x_{*},x)(K + \sigma^{2}I)^{-1}(y-m(x))
\end{equation}
and
\begin{equation}
    k_{post}(x_{*},x_{*}) = k(x_{*},x_{*}) - k(x_{*},x)(K + \sigma^{2}I)^{-1}k(x,x_{*})
\end{equation}
The $(K + \sigma^{2}I)^{-1}$ term is called the Kalman Gain\cite{welch1995introduction}, where $K$ is the kernel function of the GP prior.\\
Now using the predictive mean and variance we can calculate the posterior belief of about the function $f$,
\begin{equation}
    \mathbb{E}[f(x_{*})|x_{*},X,y] = m(x_{*}) = k(X,x_{*})^{T}(K + \sigma^{2}I)^{-1}y
\end{equation}
\begin{equation}
    \mathbb{V}[f(x_{*})|x_{*},X,y] = k(x_{*}) = k(x_{*},x_{*}) - k(x_{*},X)(K + \sigma^{2}I)^{-1}k(X,x_{*})
\end{equation}
The posterior belief for a function can also be represented graphically as:
\begin{figure}[H]
  \centering
  \includegraphics[width=0.75\textwidth]{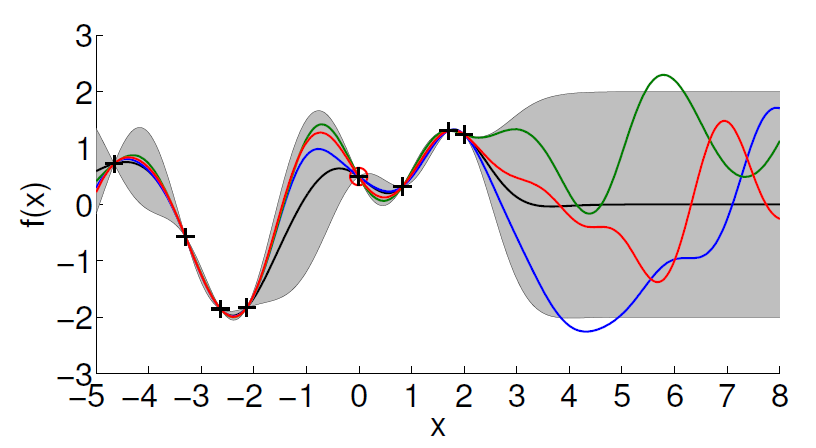}
  \caption{Posterior belief of a function}
  \label{fig:post}
\end{figure}
\section{Analyzing 100 patients with 10+ visits only}
I started the work off firstly by focusing on the main research paper which is the baseline for this idea\cite{marinescu2018tadpole} followed by studying the paper by one Ognjen's paper\cite{peterson2017personalized}. The dataset is taken from the TADPOLE challenge. So going through the dataset page of the TADPOLE challenge \href{https://tadpole.grand-challenge.org/Data/}{TADPOLE Dataset} was very imperative.\\
Around 100 patients had their CS status classified as CN, MCI or AD in the dataset. Peterson\cite{peterson2017personalized} already created the model for forecasting ADAS-Cog13 for (t+1) visit only for those 100 patients which had more than 10+ visits only and had at least 17.5\% data available.I used that idea and created the source model using auto-regressive GP\cite{candela2003propagation} for each of the group followed by personalize GP using domain adaptive GP\cite{eleftheriadis2017gaussian} for new patients for each of the group ultimately creating three source and three personalized models. I then analyzed the Mean Absolute Error\cite{willmott2005advantages} for each of the different source and personalized models for each new patient when each of the new patient can either have CN, MCI or AD as group. The models are created on the basis of multi-modal feature vector as input features. We have the ADAS-Cog13 score for each of these patients ready which are used as labels.\\
The following table depicts the the different error rates for the different groups in terms of Maximum Absolute Errors(MAE):
\begin{table}[H]
\centering
{\rowcolors{2}{lightgray}{}
 \begin{tabular}{| c | c | c | c | c |} 
 \hline
 \rowcolor{green!80!yellow!50} Model & CN & MCI & AD & Average \\ [0.5ex] 
 \hline\hline
 $S_{CN}$ & 0.0954 & 14.439	& 14.137 & 9.5572 \\ 
 \hline
 $S_{MCI}$ & 5.4696 & 1.3031 & 6.8578 & 4.5435 \\
 \hline
 $S_{AD}$ & 4.5478 & 4.2674	& 1.9746 & 3.5966 \\
 \hline
 $P_{CN}$ & 0.1029 & 13.338	& 10.963 & 8.1347 \\
 \hline
 $P_{MCI}$ & 5.3284 & 1.3134 & 5.688 & 4.1099 \\
 \hline
 $P_{AD}$ & 4.2895 & 3.7544 & 2.06 & 3.368 \\ [1ex] 
 \hline
\end{tabular}
}
\caption{The MAE rate for different source and personalized GPs for each group}
\end{table}
In the above table, the $S_{*}$ depicts the source model for each of the three different Clinical Statuses whereas the $P_{*}$ depicts the personalized models for each of the groups. The last column is the average MAE calculate for each source and personalized model.\\
Next I tried to create a clustering of the three different groups based on the ADAS-Cog13 score as feature vector. I have used k-means clustering\cite{kanungo2002efficient} to solve the issue although we can also use spectral clustering\cite{ng2002spectral} for better visualization. In Spectral clustering, the eigenvalues are used from the Graph Laplacian Matrix\cite{merris1994laplacian} which is basically the difference between degree matrix and adjacency matrix.\\
The following figure illustrates the clustering formed based on ADAS-Cog13 score for the three different groups:
\begin{figure}[H]
  \centering
  \includegraphics[width=0.75\textwidth]{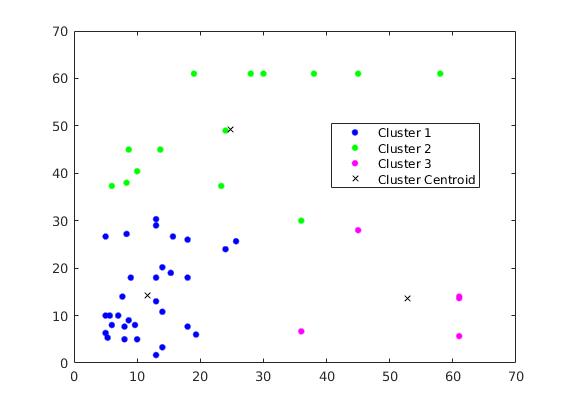}
  \caption{Three different clusters based on ADAS-Cog13 score}
  \label{fig:cluster}
\end{figure}
\begin{itemize}
    \item Cluster 1 represents CN cluster.
    \item Cluster 2 represents MCI cluster.
    \item Cluster 3 represents AD cluster.
\end{itemize}
\section{Survival Analysis}
The concept of survival analysis from Daniel's paper\cite{bello2019deep} is to be used in this project. I will be using a right-censored-time-to-event outcome\cite{faraggi1995neural} in our case just like in the paper. These output ADAS-Cog13 scores are handled using survival analysis techniques and we will be using Cox's proportional hazards regression model\cite{cox1972regression}:
\begin{equation*}
    \log \frac{h_{j}(t)}{h_{0}(t)} = \beta_{1}z_{j1}+\beta_{2}z_{j2}+...+\beta_{k}z_{jk}
\end{equation*}
where, $h_{j}(t)$ is the hazard function for subject $j$; which is the normalized probability(chance) of patient $j$ converting to AD in time $t$. We have the term $t$ set as 24 months. The term $h_{0}(t)$ is the baseline hazard level to which all subject-specific hazards $h_{j}(t)(i=1,...,n)$ are compared.\\
For us, the $h_{0}t$ is the ADAS-Cog13 score at t = $0^{th}$ month visit.
The $z_{j1}, z_{j2}...$ are the predicted feature vector for each subject $j$. The $z_{j1}$ is the 1st predicted visit for patient $j$, i.e, the (t+1) timestep of the ADAS-Cog13 score predicted. And similarly for $z_{j2}$ is the (t+2) timestep of predicted  ADAS-Cog13 score. It goes on for the 4 timesteps for each patient for all the patients. That means we will be predicting the probability for 6,8,12 and 24 months into the future for each patient.\\
The $\beta1,\beta2...$ values are the unknown weight parameters which can be estimated by Max likelihood estimation.
\chapter{Data Analysis and Preprocessing}
In this part I would provide a detailed explanation on how the dataset was analyzed and pre-processed before being fed into the GP model. The \href{https://tadpole.grand-challenge.org/Data/}{TADPOLE Dataset} has 1737 patients in total with various missing values , missing visits and other sparse data with unnecessary noise. This section deals with all those factors by first analyzing the dataset and then cleaning up the dataset before feeding it to the GP model followed by the Cox model and classifier.
\section{Data Analysis}
First we perform various forms of data analysis on the dataset to get some idea about the trends and values in the dataset. The missing values in of ADAS-Cog13 and CS columns were replaced by the previous values in those columns and all the remaining missing values of the other columns were replaced by -9999999.\\
Following replacement of the -9999999 missing values with the previous non-missing values I calculated the the general trajectory of the ADAS-Cog13 scores of the different patients in various visit times, where each patient represented by their RIDs(or patient IDs). This is shown in the figure in the folowing page:
\begin{figure}[H]
  \centering
  \includegraphics[width=0.75\textwidth]{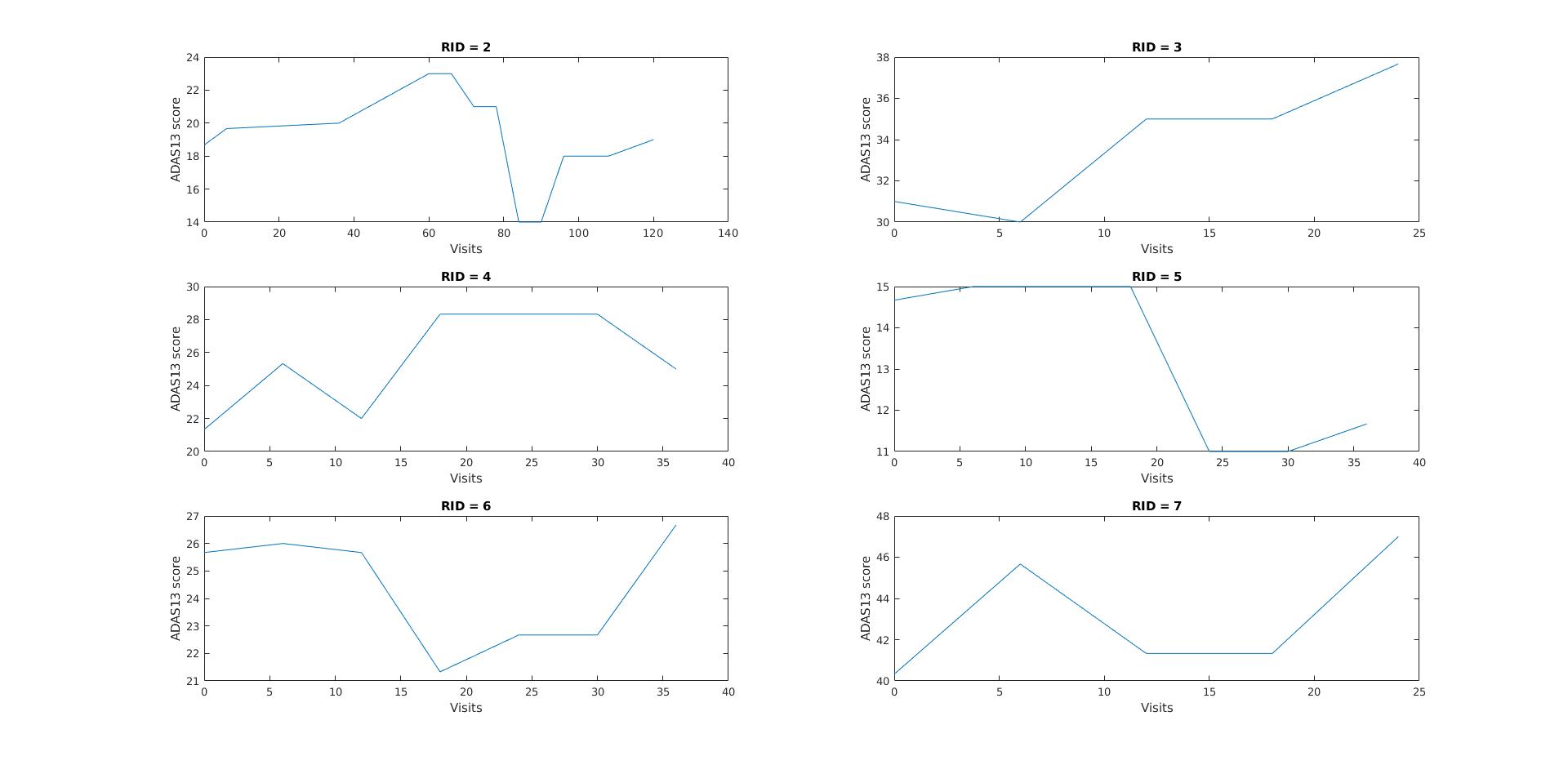}
  \caption{Trajectories for few patients based on ADAS-Cog13 scores}
  \label{fig:visitsVADAS}
\end{figure}
Next, I mapped the CN, MCI and AD groups based on ADAS-Cog13 scores in a single mapping for all the the 1737 patients. And hence, the graph which is generated is:
\begin{figure}[H]
  \centering
  \includegraphics[width=1.25\textwidth]{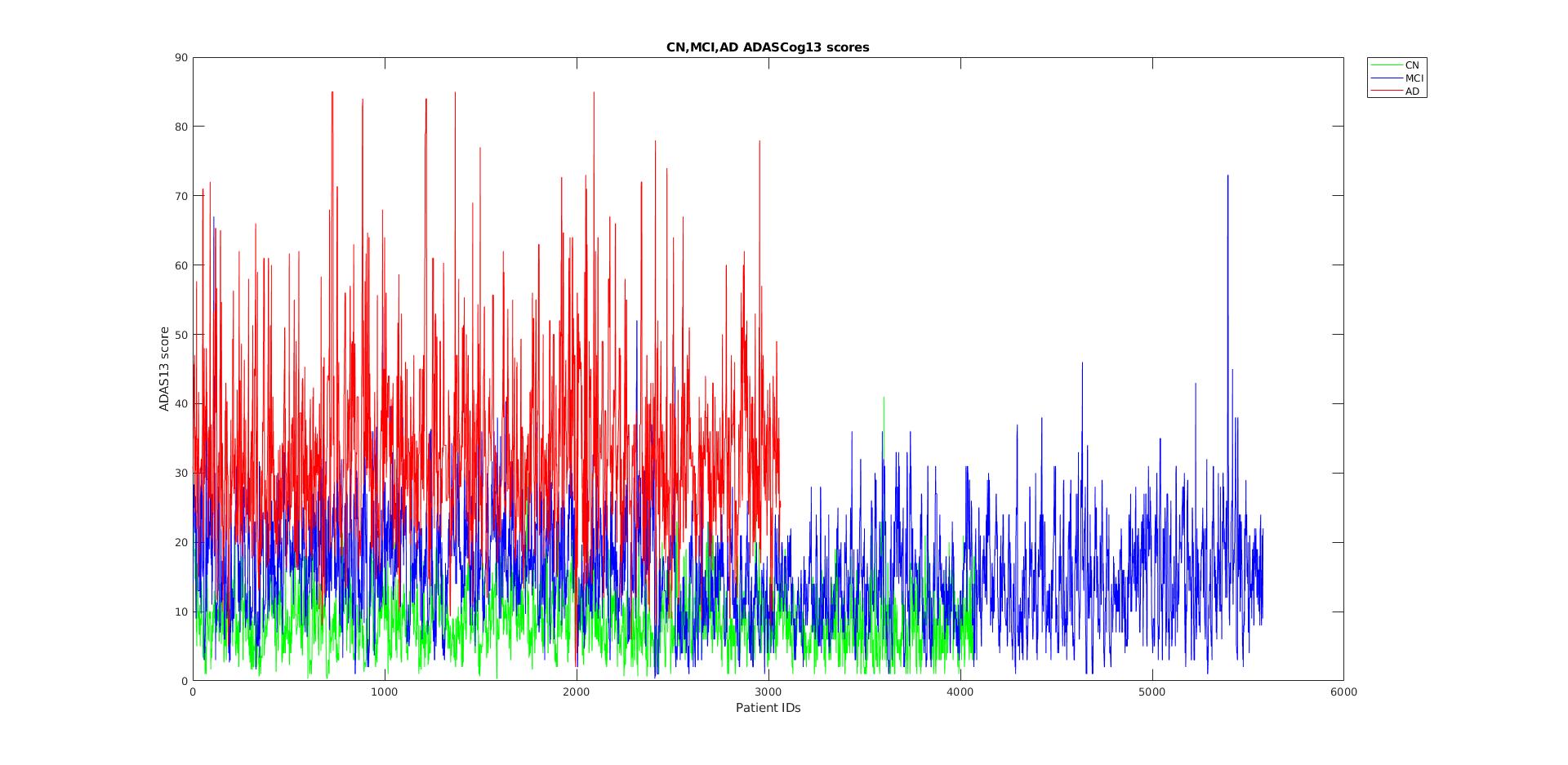}
  \caption{CN, MCI and AD categories for different ADAS-Cog13 score for each patient}
  \label{fig:CNMCIAD}
\end{figure}
As we can clearly the CN (represented by the green line) has the lowest range of ADAS-Cog13 score, followed by MCI (blue line) in the middle and the AD range (red line) representing the highest ADAS-Cog13 score. \\
Although there is a significant amount of variance in the ADAS-Cog13 scores in each group but it is quite evident that the lower the ADAS-Cog13 score the more chances that the patient falls into the CN category and higher the score higher the chances of falling into AD category.\\
After this I plotted the histograms of ADAS-Cog13 scores per group for each of the three groups.
\begin{figure}[H]
  \centering
  \includegraphics[width=0.90\textwidth]{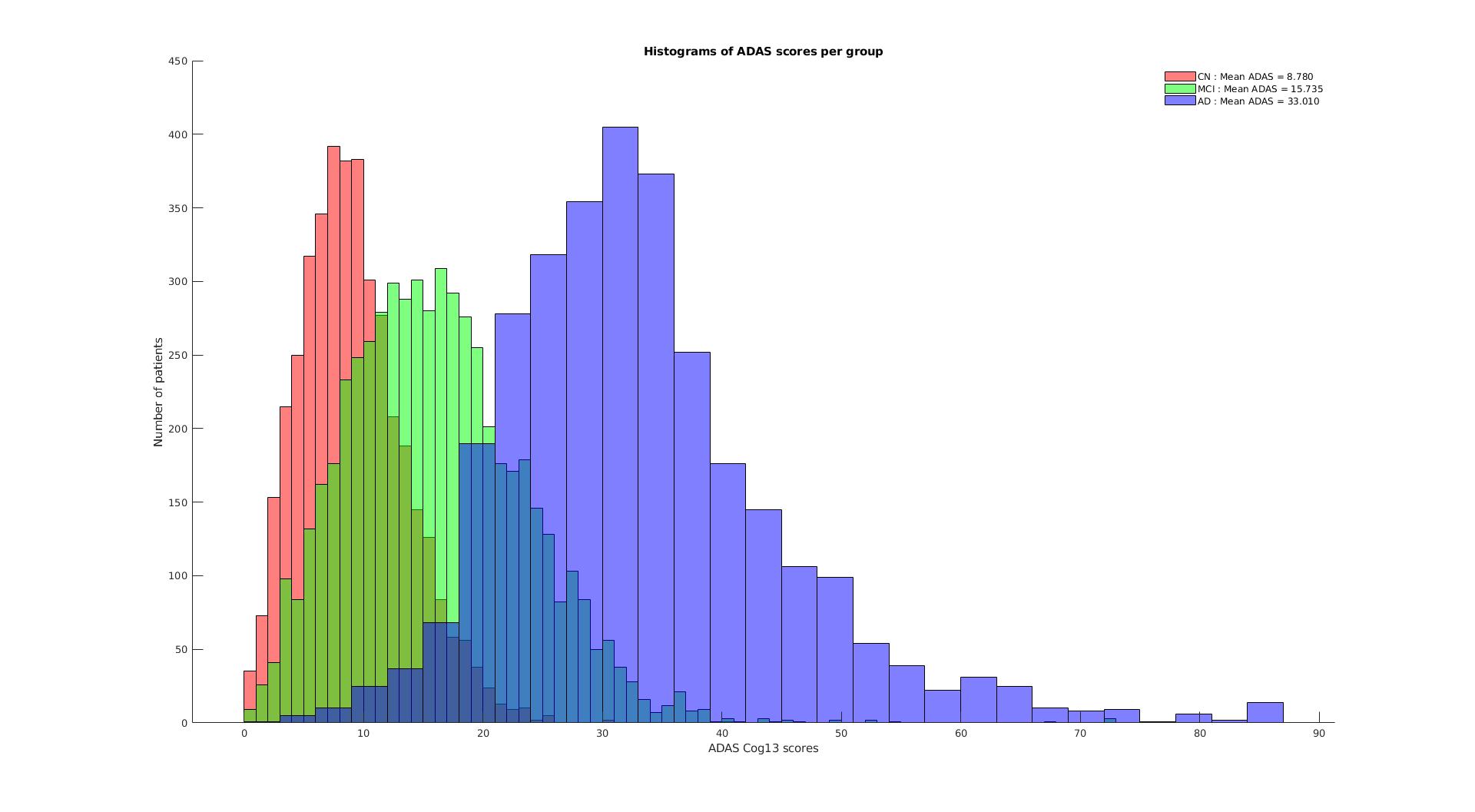}
  \caption{Histograms of ADAS scores per group}
  \label{fig:Hist_CNMCIAD}
\end{figure}
The red colored histograms depict the CN group ADAS-Cog13 scores. The green colored histograms depict the MCI group ADAS-Cog13 scores. The blue colored histograms depict the AD group ADAS-cog13 scores.\\
\textbf{Mean ADAS-Cog13 of CN group = 8.780}\\
\textbf{Mean ADAS-Cog13 of MCI group = 15.735}\\
\textbf{Mean ADAS-Cog13 of AD group = 33.010}\\
\textbf{Standard Deviation ADAS-Cog13 of CN group = 4.4512}\\
\textbf{Standard Deviation ADAS-Cog13 of MCI group = 7.5023}\\
\textbf{Standard Deviation ADAS-Cog13 of AD group = 11.7477}
\subsection{Table for the number of patients in each group}
\begin{figure}[H]
  \centering
  \includegraphics[width=0.75\textwidth]{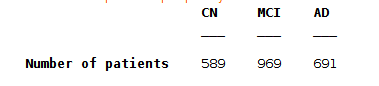}
  \caption{Number of patients per group}
  \label{fig:No.p}
\end{figure}
There are a total number of 1737 patients. In the above table we see the patients being divided into the three groups: CN, MCI and AD. Some patients fall under the category of multiple groups as there is change in group,i.e. a conversion during the patients' visit period.
\subsection{Plot trajectories for each subgroup CN, MCI, AD, i.e, Plot Average trajectory (Along Time) per group + Standard Deviation}
\begin{figure}[H]
  \centering
  \includegraphics[width=1.00\textwidth]{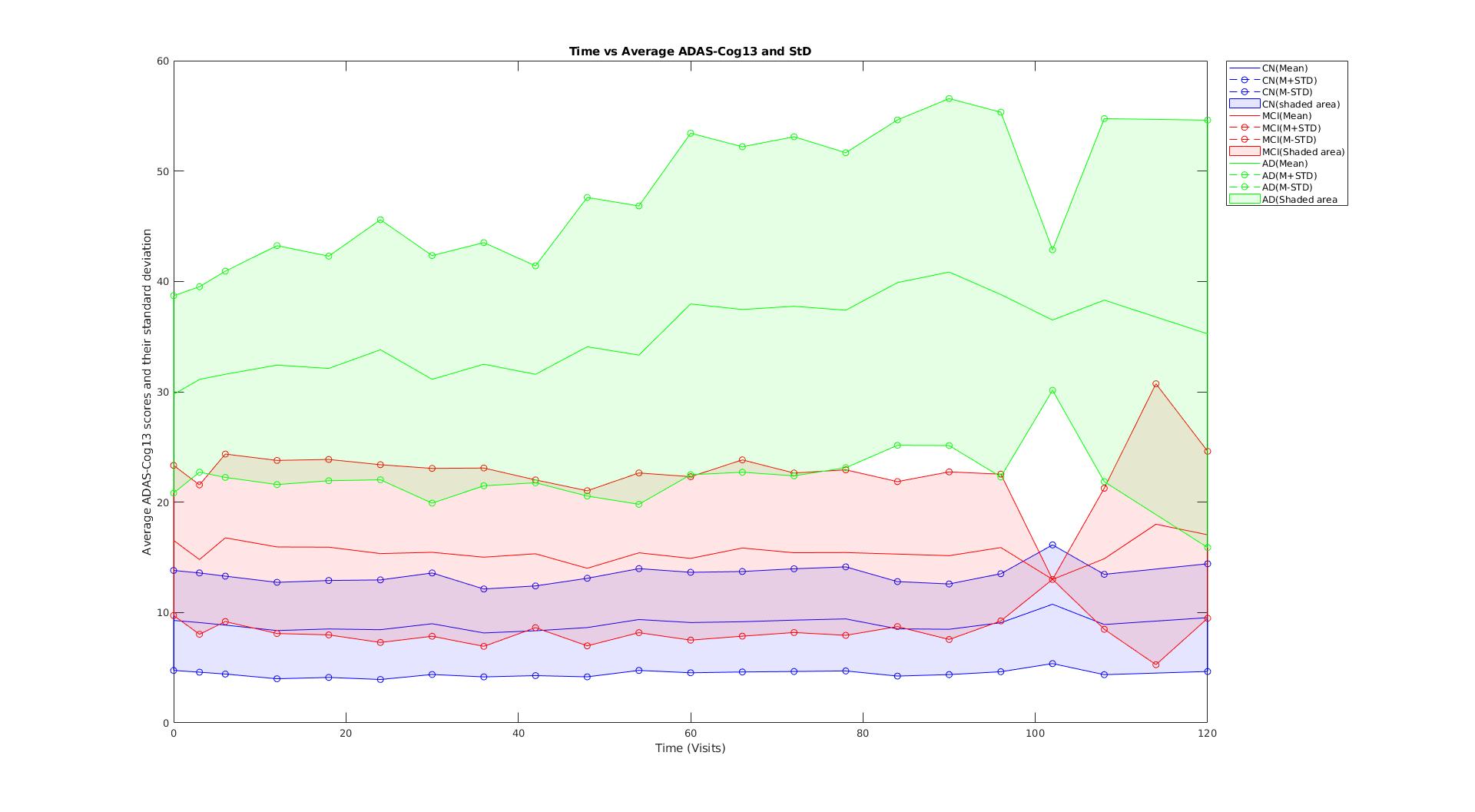}
  \caption{Time vs Average ADAS-Cog13 and Standard Deviation}
  \label{fig:No.p}
\end{figure}
The blue shaded region is the region of CN Mean $\pm$ Standard deviation. The red shaded region is the region of MCI Mean $\pm$ Standard deviation. The green shaded region is the region of AD Mean $\pm$ Standard deviation. The solid line is the average trajectory in each group. The dotted lines are the $\pm$ standard deviation lines.
\subsection{Analysis of the change in ADAS-Cog13 score in the 4 sliding windows}
Here we calculate the mean, the maximum , the median, the minimum and the standard deviation of the difference in ADAS-Cog13 scores for each patient for a span of 0-2 years. These 0-2 years is split among 4 sliding windows where each window consists of 6 months.\\
\\
\textbf{For each window of 4 steps},\\
\\
\textbf{1st Window (0-6 months)}:\\
Mean of difference in ADAS-Cog13 score = 0.4487\\
Max of difference in ADAS-Cog13 score = 25\\
Min of difference in ADAS-Cog13 score = -46.6700\\
Median of difference in ADAS-Cog13 score = 0\\
Standard deviation of difference in ADAS-Cog13 score = 4.7573\\
\\
\textbf{2nd Window (6-12 months)}:\\
Mean of difference in ADAS-Cog13 score = 0.2893\\
Max of difference in ADAS-Cog13 score = 48\\
Min of difference in ADAS-Cog13 score = -42\\
Median of difference in ADAS-Cog13 score = 0\\
Standard deviation of difference in ADAS-Cog13 score = 4.8433\\
\\
\textbf{3rd Window (12-18 months)}:\\
Mean of difference in ADAS-Cog13 score = 0.3295\\
Max of difference in ADAS-Cog13 score = 14.3300\\
Min of difference in ADAS-Cog13 score = -16.0000\\
Median of difference in ADAS-Cog13 score = 0\\
Standard deviation of difference in ADAS-Cog13 score = 2.4348\\
\\
\textbf{4th Window (18-24 months)}:\\
Mean of difference in ADAS-Cog13 score = 1.6002\\
Max of difference in ADAS-Cog13 score = 33.6700\\
Min of difference in ADAS-Cog13 score = -14.3400\\
Median of difference in ADAS-Cog13 score =  1\\
Standard deviation of difference in ADAS-Cog13 score = 5.1205\\
\\
The calculation of difference in ADAS-Cog13 score is important as this will be later used to map against the change in CS values. The feature set will be change in ADAS-Cog13 scores for each patient against the label of change in CS values.
\section{Data Preprocessing}
\subsection{Histogram of the number of visits visits for each patient}
\begin{figure}[H]
  \centering
  \includegraphics[width=1.25\textwidth]{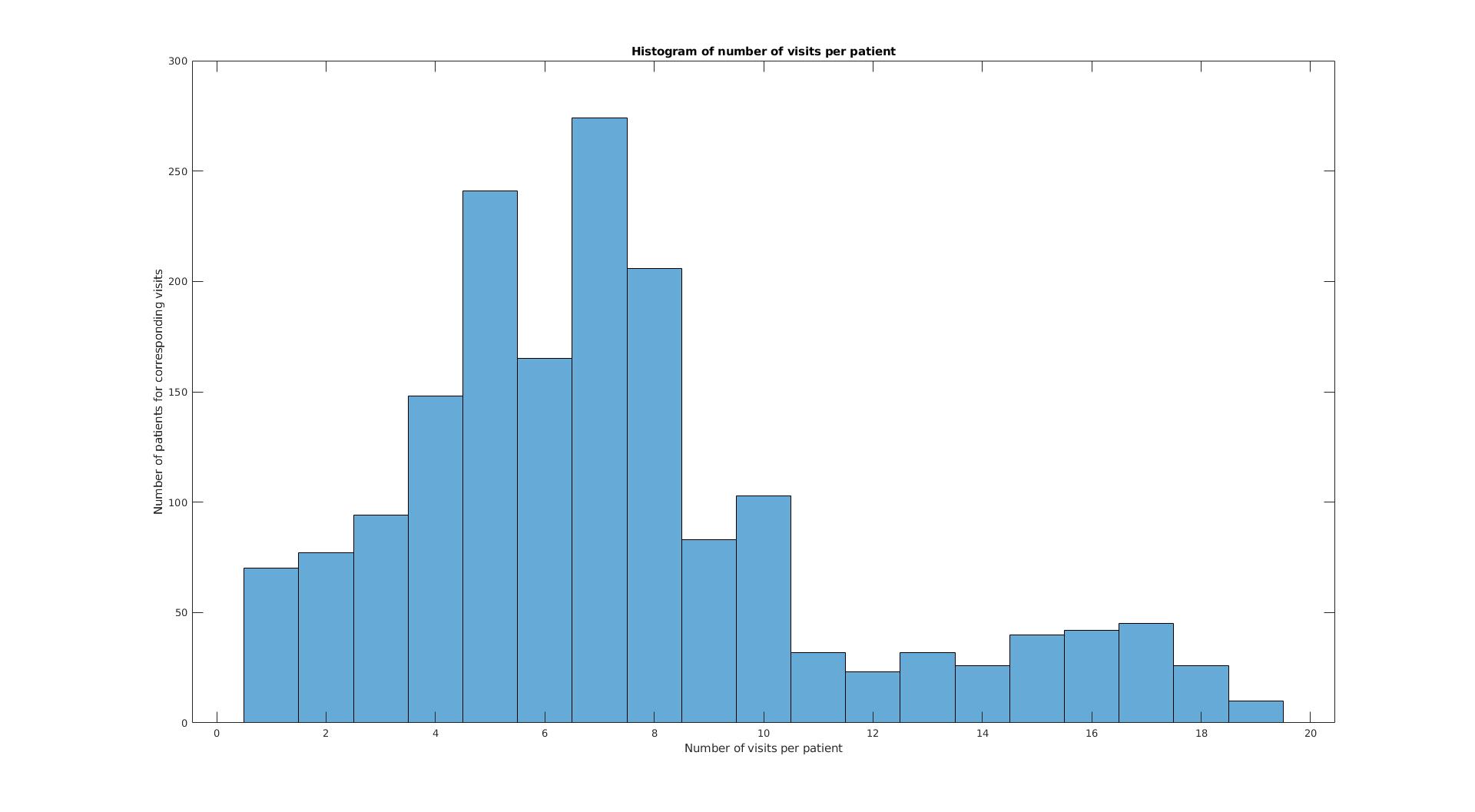}
  \caption{Histogram of number of visits per patient}
  \label{fig:No.p}
\end{figure}
Here we clearly identity which patients have less than 4 visits. Its easier to remove those patients from this histogram.\\
There are a total of \textbf{241} patients with less than 4 visits.\\
\textbf{Patient IDs of those patients}: $38,44,45,103,111,168,182,219,228,243,245,284,\\
288,319,332,339,354,356,360,397,405,407,409,417,436,438,442,445,446,\\
469,484,489,492,506,590,592,613,640,642,653,691,699,721,\\
743,761,777,816,825,828,841,853,860,871,876,880,884,890,\\
917,924,929,955,957,958,1013,1015,1021,1033,1037,1051,1083,\\
1104,1113,1137,1149,1154,1184,1185,1191,1192,1199,1204,1210,\\
1231,1244,1248,1257,1263,1277,1294,1306,1322,1334,1337,1338,\\
1340,1350,1354,1357,1363,1366,1391,1400,1409,1411,1412,1420,\\
1423,1426,1435,2003,2011,2057,2070,2138,2171,2193,2194,2199,\\
2237,2278,2351,4066,4085,4095,4097,4139,4186,4201,4209,4251,\\
4260,4264,4274,4332,4348,4353,4362,4403,4408,4459,4474,4524,\\
4540,4546,4564,4575,4583,4601,4603,4609,4622,4633,4688,4692,\\
4694,4708,4719,4740,4745,4792,4793,4798,4803,4859,4907,4914,\\
4924,4943,4963,4968,4976,4980,4990,4994,4997,5005,5016,5032,\\
5059,5075,5090,5091,5102,5112,5120,5138,5146,5148,5150,5160,\\
5171,5184,5195,5204,5205,5206,5208,5209,5210,5212,5213,5218,\\
5222,5224,5227,5228,5231,5240,5241,5242,5244,5248,5250,5251,\\
5252,5253,5256,5259,5261,5262,5265,5266,5267,5271,5272,5275,\\
5277,5278,5279,5280,5282,5283,5285,5287,5288,5289,5290,5292,\\
5294,5295,5296.$\\
So we can clearly remove these patients while performing calculations because they have less than 4 visits, i.e, not enough data to predict AD conversion for 6,12,18 and 24 months.\\
Hence, now we have 1737-241 = \textbf{1496} patients with at least 4 visits.
Next we select patients with 5 or 5+ visits with the 0,6,12,18 and 24 months visits available. We need all these 5 visits at the minimum to calculate whether a patient converts within 2 years or not. If any visit is missing from the first 24 months then it will not be much value addition to our training dataset. So, we ignore those patients who do not have 0,6,12,18 and 24 months in their visit history. There were quite a few patients with $3^{rd}$ month visit as well; we included those patients as well but did not include the visit number 3 in calculations as it would make the 4 sliding windows uneven.\\
So, from 1496 patients we now select a total of \textbf{1141} patients to work with.
\subsection{Histogram of the number of visits visits for 1141 patients}
\begin{figure}[H]
  \centering
  \includegraphics[width=1.05\textwidth]{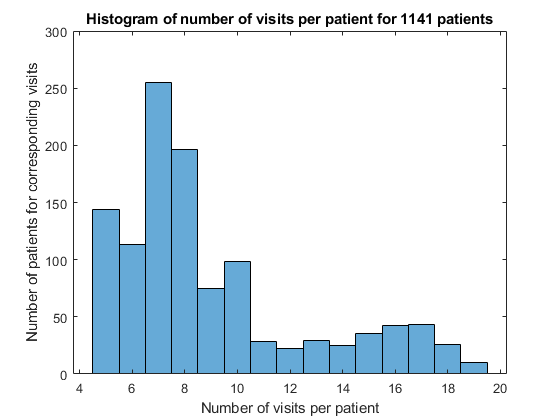}
  \caption{Histogram of number of visits for 1141 patients}
  \label{fig:1141p}
\end{figure}
The figure \ref{fig:1141p} represents the number of visits per patient for the 1141 patients that has 6,12,18 and 24 months in their visit list.
\subsection{Histogram for percentage of missing data in 1141 patients}
\begin{figure}[H]
  \centering
  \includegraphics[width=1.15\textwidth]{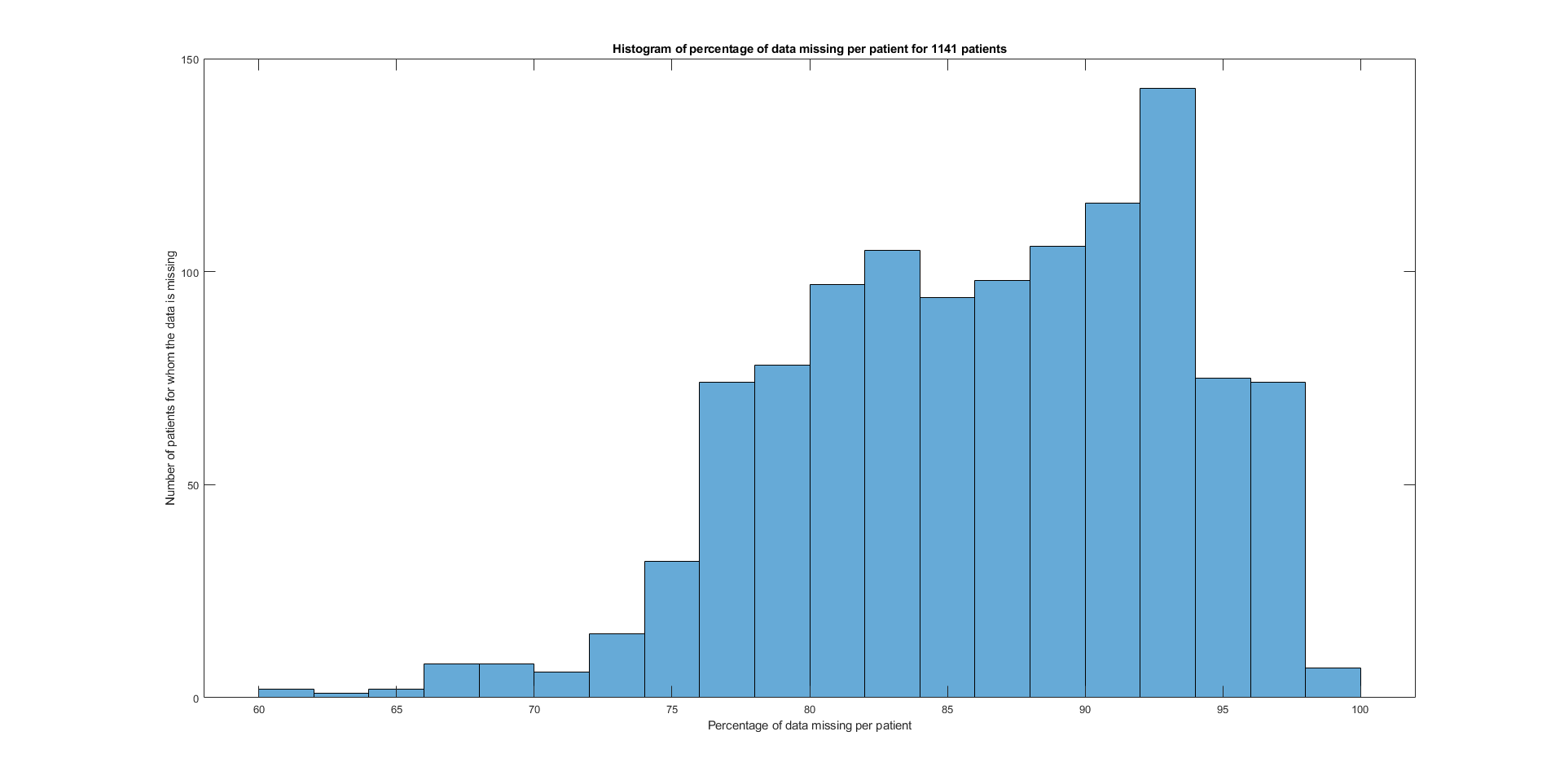}
  \caption{Histogram of number of visits for 1141 patients}
  \label{fig:Missing}
\end{figure}
From figure \ref{fig:Missing} we deduce that around 415 patients are such whose more than 90\% data is missing. Thus 415 patients are such that at most 10\% is also not available for them; so we will exclude them.\\
Hence, finally the number of patients to work with is 1141-415 = \textbf{726} patients.
\subsection{Selecting the features}
Initially there were 12739 rows in the tadpole dataset file for 1737 patients. Now, there are 6584 rows for 726 patients that we have chosen to work with. Also, there are around 1465 columns available in the feature set but we will try to select around 616 columns as feature set as in Rudovic's initial paper \cite{peterson2017personalized}.\\
\begin{itemize}
    \item Cognitive Tests: We chose 9 features from the Cognitive Tests namely :- ADAS-Cog13, ADAS-Cog11, MMSE, CDSRB, RAVLT-immediate subtype, RAVLT-percent forgetting subtype, RAVLT-forgetting subtype, FAQ and RAVLT-learning subtype.
    \item MRI or Magnetic Resonance Imaging: We chose around 366 MRI biomarkers which included the three main MRI markers:- ROI cortical thickness, ROI volume and ROI surface areas.
    \item DTI or Diffusion Tensor Imaging : We chose 229 DTI features for the patients based on mainly three types of DTI measurements:- mean diffusivity, radial diffusivity and axial diffusivity.
    \item Genetics: We chose 3 features from genetics which are the APOE E4, APGEN1 and APGEN2 columns. The APOE E4 is a huge detector whether a patient will get Alzheimer's or not. The APGEN1 and APGEN2 are two types of APOE E4 based on alleles.
    \item Demographics: We chose 6 features from Demographics namely :- gender, age, race, ethnicity, years of education and marital status.
    \item CSF or Cerebrospinal Fluid: We chose again 3 features from CSF namely :- amyloid beta, phosphorylated tau and tau. They are important of early detection of dementia in a patient.
\end{itemize}
So, finally we have 616 features to work with and 726 patients. The CS, Patient ID columns are added to the dataset but not used as a column. Also, the ADAS-Cog13 score used as a label in the case of the GP model and not as a feature. The rest 615 columns are used as input features.\\
We then set the dataset in such a way that first are 615 features followed by the last 4 columns as 6,12,18 and 24 months ADAS-Cog13 values which are to be used as labels for each visit of a patient, i.e, the (t+1),(t+2),(t+3) and (t+4) ground truth values; which is shown in the next page.
\begin{figure}[H]
  \centering
  \includegraphics[width=0.35\textwidth]{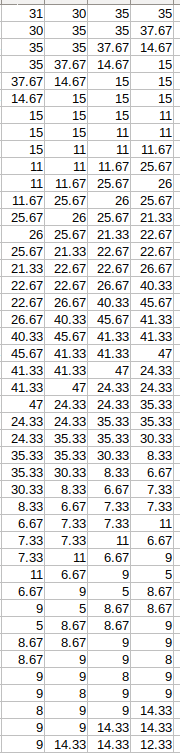}
  \caption{6,12,18 and 24 months ADAS-Cog13 ground truth}
  \label{fig:Ground}
\end{figure}
\chapter{Forecasting of ADAS-Cog13 score}
\section{Z-score normalization}
The 726 patients along with the 616 input features are stored in a csv file. Except the ADAS-Cog13 scores all the remaining 615 features are z-normalized\cite{cheadle2003analysis}. In z-normalization each element in the feature matrix is centred to have mean zero and a scaled standard deviation of 1 which is done column wise.

\section{Gaussian Process setting}
Let us consider a supervised setting which has $I^{(k)} = (i_{a_{k}}^{(k)})_{a_{k}=1}^{A_{k}}$, where $I^{(k)}$ represents the multi-modal feature vector which we use as input features of $A_{k}$ patients. The output score is the predicted ADAS-Cog13 $\in$ $(0-85)$ which is saved as $J^{(k)} = (j_{a_{k}}^{(k)})_{a_{k}=1}^{A_{k}}$. Now, each patient is represented as a pair of data: ($i_{a_{k}}^{(k)},j_{a_{k}}^{(k)}$). This data pair is represented such that $i_{a_{k}}^{(k)} = (i_{1},...,i_{t})$ which has the input features up to $t^{th}$ visit and we have the corresponding output ADAS-Cog13 scores $j_{a_{k}}^{(k)} = (j_{p})$. In our case we have p = 6,12,18 and 24 months i.e, p = t+1,t+2,t+3 and t+4. These data pairs for each patient are used as training the forecasting model. From here, we calculate the source GP, the personalized (or domain adaptive GP) and the target GP\cite{peterson2017personalized}.
\section{Training/Testing of the GP models}
We perform a 10-fold cross validation which is independent of patients while training and testing the GP model. After that the we calculate the Mean Absolute Error (MAE) of each fold and report the mean($\pm$ standard deviation) of all the folds together.
The following are the reported MAE Error for the 10 folds for sGP, pGP and tGP model:
\begin{figure}[H]
  \centering
  \includegraphics[width=0.85\textwidth]{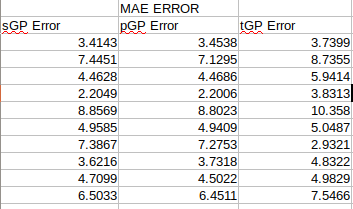}
  \caption{MAE Error for the 10-folds}
  \label{fig:MAE10}
\end{figure}
The mean($\pm$ standard deviation) MAE error comes as:
\begin{table}[H]
\centering
\begin{tabular}{|| c | c | c | c ||} 
 \hline
 Error & sGP & pGP & tGP \\ [0.5ex] 
 \hline
   & 5.35$\pm$2.11 & 5.29$\pm$2.05 & 5.79$\pm$2.38 \\ [1ex] 
 \hline
\end{tabular}
\caption{The mean($\pm$SD) MAE rate for 10-folds}
\end{table}
Now, we report the forecasted ADAS-Cog13 for 6,12,18 and 24 months which are predicted as the output of the GP model as mean and covariance function, i.e, GP(m,k) where m is the mean function and k is kernel or covariance function. This is done for each of the three models, i.e, the source model(sGP), the personalized model(pGP) and the target model(tGP).
\begin{figure}[H]
  \centering
  \includegraphics[width=0.75\textwidth]{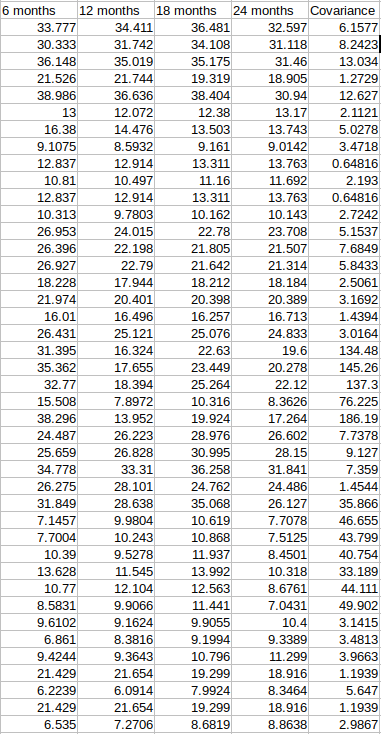}
  \caption{sGP model prediction for 6,12,18 and 24 months}
  \label{fig:sGP}
\end{figure}
\begin{figure}[H]
  \centering
  \includegraphics[width=0.75\textwidth]{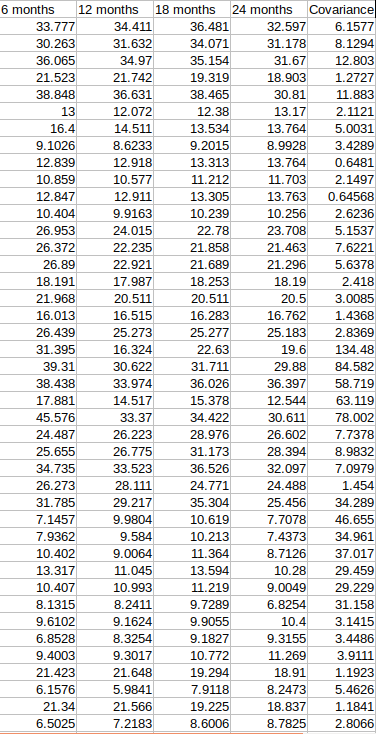}
  \caption{pGP model prediction for 6,12,18 and 24 months}
  \label{fig:pGP}
\end{figure}
\begin{figure}[H]
  \centering
  \includegraphics[width=0.75\textwidth]{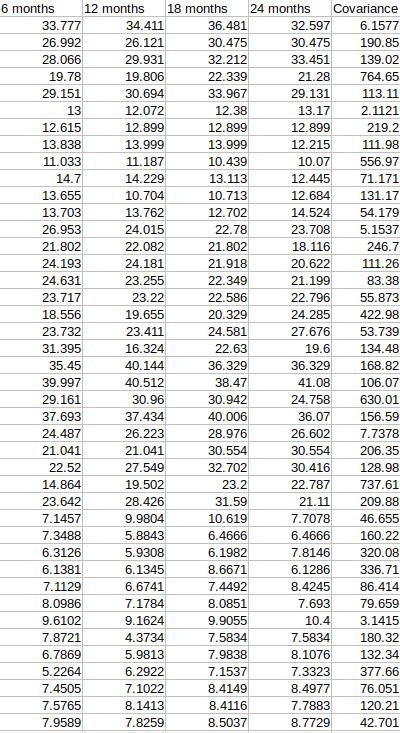}
  \caption{tGP model prediction for 6,12,18 and 24 months}
  \label{fig:tGP}
\end{figure}
\section{Comparison}
AS we can clear see in terms of MAE Error or by plainly comparing it to the ground truth the pGP models performs the best among the three models. This is the advantage of the domain adaptive Auto-regressive GP. This property was clearly demonstrated in the paper \cite{peterson2017personalized}. But through the dataset which we had and performed sGP,pGP and tGP on that dataset as well; pGP turned yet again to be the best model amongst the three models.\\
In the below graph we show the different MAE rate comparison for the three models for 10-folds:
\begin{figure}[H]
  \includegraphics[width=1.25\textwidth]{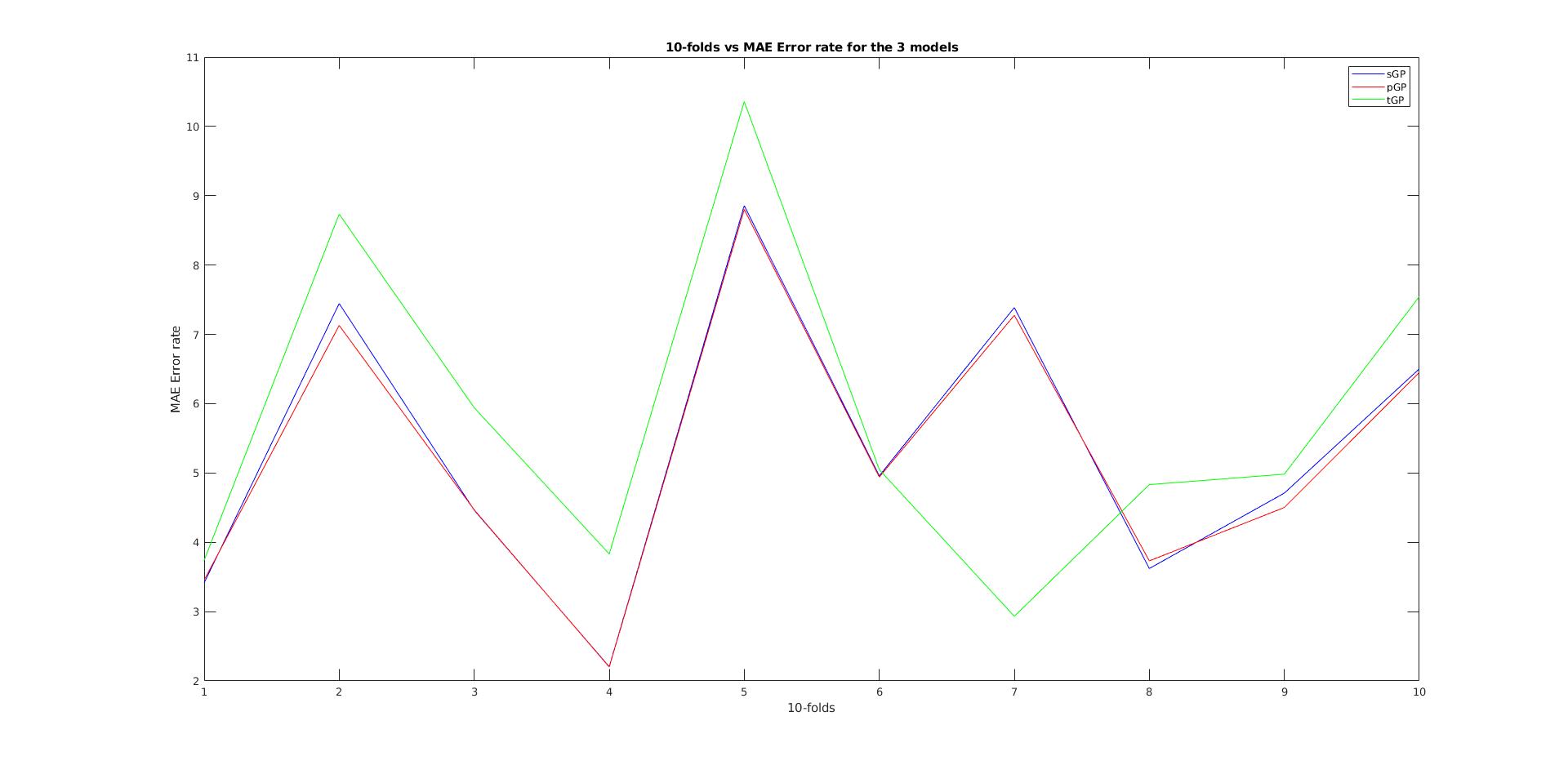}
  \caption{MAE Error comparison for the three models}
  \label{fig:compare}
\end{figure}
The blue line is the sGP model, the red is the pGP model and the green is the tGP model. As we can see, the pGP model has the least error rates in nearly all folds except the $9^{th}$ fold where sGP has the lowest error rates and also the $7^{th}$ fold where tGP has the lowest error rate. In general, tGP performs the worst and pGP performs the best.
\chapter{Predicting Conversion in Patients}
In this part we use the forecasted ADAS-Cog13 scores for 6,12,18 and 24 months and the ground truth CS values of the patients to check whether a patient converts within 24 months or not. We will be using Cox model initially to predict the normalized probability of conversion of patient within the 4 different windows of 6,12,18 and 24 months. And finally an SVM(Support Vector Machines)\cite{suykens1999least} classifier to predict whether the patient actually converts or not.
\section{Setting up the dataset}
First, we take the average of all the 6,12,18 and 24 months predictions of ADAS-Cog13 scores. That is, the average of 6 months scores for sGP, pGP and tGP , the average of 12 months scores for sGP, pGP and tGP etc. The result is 4 column matrix with the average ADAS-Cog13 scores of sGP,pGP and tGP for (t+1),(t+2),(t+3) and (t+4) time period. This 4 columns are placed beside the Patient ID and Clinical Status columns and overall these 6 columns are stored as a csv file which is shown in the next page.
\begin{figure}[H]
  \centering
  \includegraphics[width=0.85\textwidth]{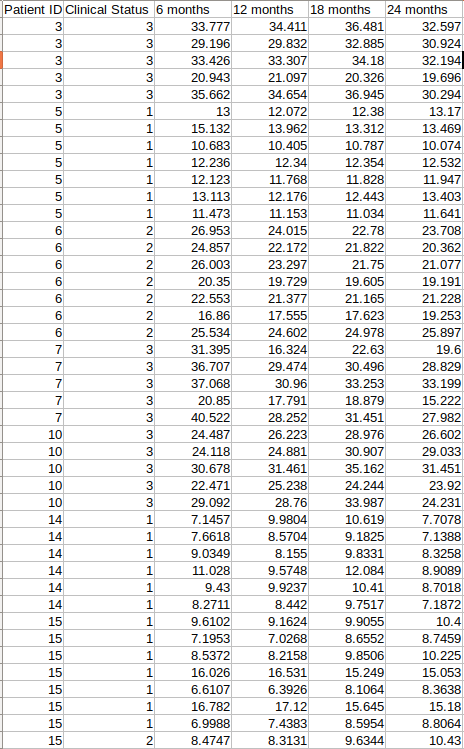}
  \caption{6,12,18 and 24 months average ADAS-Cog13 score along with Clinical Status}
  \label{fig:set}
\end{figure}
\section{Cox model}
After this we calculate the difference in ADAS-Cog13 score and the change in Clinical Status. The change is clinical status column is represented as (0/1). If there is a conversion then it is set as 1 else it is set as 0. This dataset is then fed into the Cox model\cite{cox1972regression}. The ADAS-Cog13 scores of four columns are set as the input variables and the change in CS column is set as a predictor variable. In MATLAB we coded the Cox model, where we need an input matrix X and a predictor vector T. Here we store ADAS-Cog13 as X which is an n-by-p matrix where p is 4 here because 4 columns and n denotes the number of patients. The change in CS column is the T predictor vector which is n-by-1 vector where n denotes the number of patients. \\
The Cox model predicts out the p-by-1 vector, i.e, the 4 probabilities of conversion for each patient. We repeat that for all the n patients to get the 4 probabilities of conversion for all the patients.\\
We get all the stats from the Cox model as below:
\begin{figure}[H]
  \centering
  \includegraphics[width=0.95\textwidth]{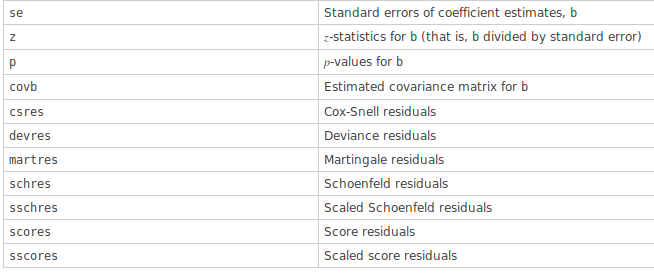}
  \caption{Values returned by the Cox model}
  \label{fig:set}
\end{figure}
We are going to use the p-values from the stats vector as this returns the normalized probability of the patients' converting chance. In the figure next page we see the normalized probability of conversion of all the patients in 6,12,18 and 24 months along with the patient ID and the ground truth change in Clinical Status.
\begin{figure}[H]
  \centering
  \includegraphics[width=0.85\textwidth]{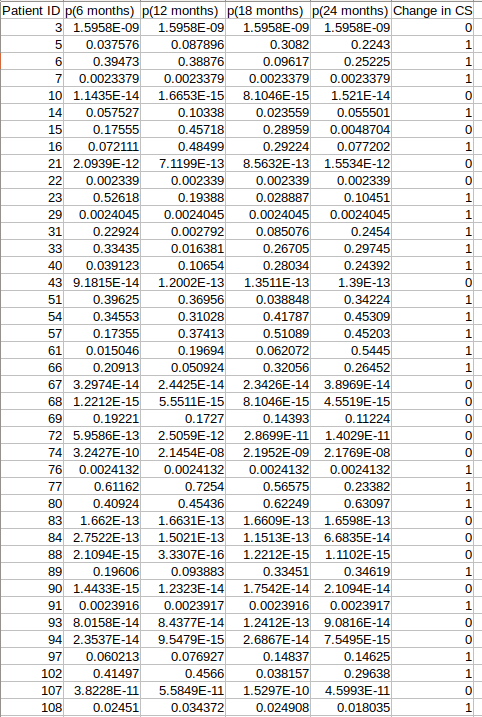}
  \caption{6,12,18 and 24 months normalized probabilities for conversion}
  \label{fig:Cox}
\end{figure}
The above figure is output of the Cox model showing the normalized probabilities of conversion for 6,12,18 and 24 months for each patient.
\pagebreak
\section{Classify whether a patient converts or not}
In this part we classify based on the ground truth of change in Clinical Statuses that whether a patient converts within 24 months or not. We will be using an SVM classifier to perform the classification. The table shown in figure \ref{fig:Cox} is used for classification. In the classifier we use the 4 normalized probability columns, i.e, n-by-4 matrix as the X value or input observations where n is the number of patients. And we use the n-by-1 change in CS column as the ground truth labels for the classifier where n is the number of patients.\\
We calculate the Precision, Recall, F1 Score and Accuracy of the classifier and report it as below:
\begin{table}[H]
\centering
\begin{tabular}{|| c | c | c | c ||} 
 \hline
 Precision & Recall & F1 Score & Accuracy \\ [0.5ex] 
 \hline
 0.7776 & 0.7914 & 0.7845 & 0.8000 \\ [1ex] 
 \hline
\end{tabular}
\caption{The different error metrics for the classifier}
\end{table}
If we consider true positives, true negatives, false positives and false negatives from the confusion matrix; then we can define precision, recall, F1 score and accuracy as:
\begin{equation}
    \begin{split}
        Precision = \frac{True Positive}{True Positive + False Positive}
        \\
        Recall = \frac{True Positive}{True Positive + False Negative}
        \\
        F1 Score = \frac{2*True Positive}{2*True Positive + False Positive + False Negative}
        \\
        Accuracy = \frac{True Positive + True Negative}{True Positive + False Positive + True Negative + False Negative}
    \end{split}
\end{equation}
As we can see we get a moderately highly accuracy of 0.8 which means around 80\% of patients were correctly classified whether they will convert within 2 years or not.\\
In next page we see the results of the classifier.
\begin{figure}[H]
  \centering
  \includegraphics[width=1.05\textwidth]{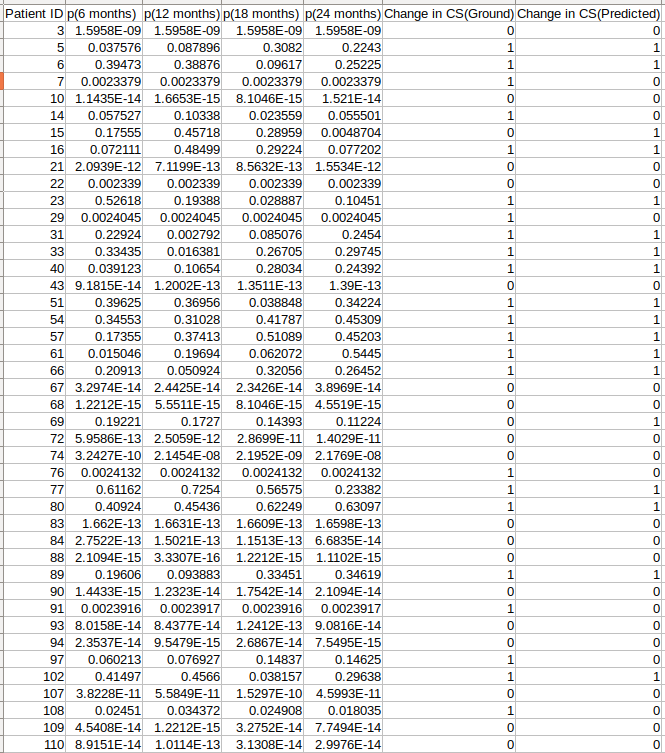}
  \caption{Classifier prediction whether each patient will convert or not}
  \label{fig:classify}
\end{figure}
The above figure clearly shows the ground truth of conversion in CS for each patient alongside the predicted conversion in CS from the classifier. This is the final part of the project which predicts for any random patient that he or she will convert or not within the span of 2 years.

\chapter{Conclusion}
In conclusion, I would like to say that the entire project was completed to help better predict the chances of a patient converting to AD. This result can help doctors a lot in understanding whether a patient actually converts to AD or not within the span of two years. This in turn can result into huge project in the domain of ML for healthcare.\\
There are a few areas where future improvements can be made. We can fill in the missing data more efficiently for the entire dataset which can thus help us predict a better value for ADAS-Cog13 scores as a lot of values are missing in the TADPOLE dataset.
Also, we have used RBF-ISO kernel in our GP models which can be replaced by ARD kernel as well and a comparison can be drawn as to which kernel produces better results. The classifier we used was SVM but there are many other classifiers which can be used for the classification whether a patient converts or not like Decision Tree classifier\cite{safavian1991survey}, Logistic regressor\cite{kleinbaum2002logistic} etc. We can also further extend the approach to say 3 years instead of 2 years and see how the exisiting GP models perform in them.
\begin{appendices}
\chapter{Appendix A}
The Ethics Checklist is shown as below:
\begin{figure}[H]
  \centering
  \includegraphics[width=0.65\textwidth]{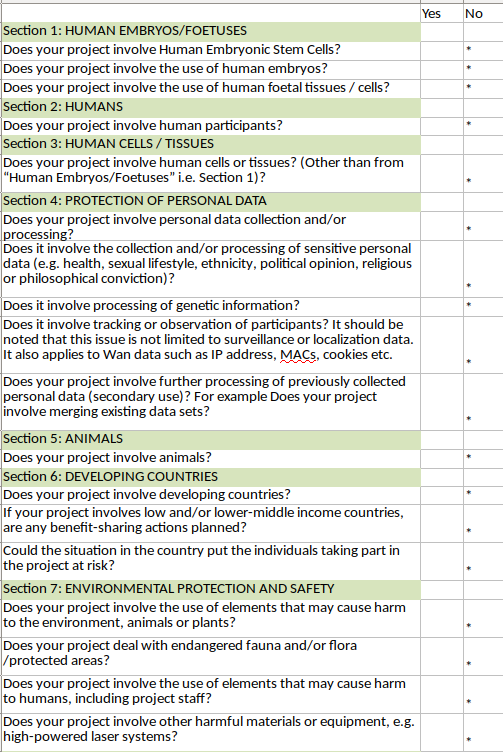}
  \caption{Ethics Checklist}
  \label{fig:ethics}
\end{figure}
\begin{figure}[H]
  \centering
  \includegraphics[width=0.65\textwidth]{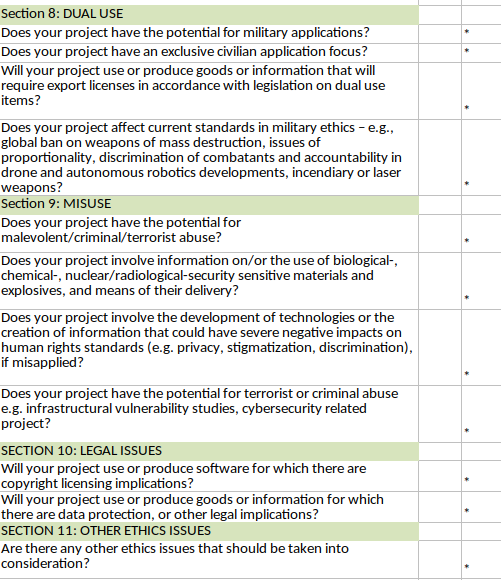}
  \caption{Ethics Checklist Continued}
  \label{fig:ethics}
\end{figure}
The above mentioned ethics checklist clearly shows that there is no human, animal or military involvement or other dual or legal issues and neither any environmental protection and safety issues are tampered with in this project. Our project did not directly involve any humans. The dataset which we took from had data collection from humans which was done by University College London for their TADPOLE dataset. We just directly took the dataset from the TADPOLE website with no direct interaction with any human for data collection.
\end{appendices}
\bibliographystyle{./bibs/IEEEtran}
\bibliography{./bibs/IEEEabrv,./bibs/sample}

\begin{thebibliography}{10}
\providecommand{\url}[1]{#1}
\csname url@samestyle\endcsname
\providecommand{\newblock}{\relax}
\providecommand{\bibinfo}[2]{#2}
\providecommand{\BIBentrySTDinterwordspacing}{\spaceskip=0pt\relax}
\providecommand{\BIBentryALTinterwordstretchfactor}{4}
\providecommand{\BIBentryALTinterwordspacing}{\spaceskip=\fontdimen2\font plus
\BIBentryALTinterwordstretchfactor\fontdimen3\font minus
  \fontdimen4\font\relax}
\providecommand{\BIBforeignlanguage}[2]{{%
\expandafter\ifx\csname l@#1\endcsname\relax
\typeout{** WARNING: IEEEtran.bst: No hyphenation pattern has been}%
\typeout{** loaded for the language `#1'. Using the pattern for}%
\typeout{** the default language instead.}%
\else
\language=\csname l@#1\endcsname
\fi
#2}}
\providecommand{\BIBdecl}{\relax}
\BIBdecl

\bibitem{marinescu2018tadpole}
R.~V. Marinescu, N.~P. Oxtoby, A.~L. Young, E.~E. Bron, A.~W. Toga, M.~W.
  Weiner, F.~Barkhof, N.~C. Fox, S.~Klein, D.~C. Alexander \emph{et~al.},
  ``Tadpole challenge: Prediction of longitudinal evolution in alzheimer's
  disease,'' \emph{arXiv preprint arXiv:1805.03909}, 2018.

\bibitem{burns2009alzheimer}
A.~Burns and S.~Iliffe, ``Alzheimer’s disease. bmj 338, b158,'' 2009.

\bibitem{cummings2006challenges}
J.~L. Cummings, ``Challenges to demonstrating disease-modifying effects in
  alzheimer’s disease clinical trials,'' \emph{Alzheimer's \& Dementia},
  vol.~2, no.~4, pp. 263--271, 2006.

\bibitem{peterson2017personalized}
K.~Peterson, O.~Rudovic, R.~Guerrero, and R.~W. Picard, ``Personalized gaussian
  processes for future prediction of alzheimer's disease progression,''
  \emph{arXiv preprint arXiv:1712.00181}, 2017.

\bibitem{folstein1975mini}
M.~F. Folstein, S.~E. Folstein, and P.~R. McHugh, ``“mini-mental state”: a
  practical method for grading the cognitive state of patients for the
  clinician,'' \emph{Journal of psychiatric research}, vol.~12, no.~3, pp.
  189--198, 1975.

\bibitem{hughes1982new}
C.~P. Hughes, L.~Berg, W.~Danziger, L.~A. Coben, and R.~L. Martin, ``A new
  clinical scale for the staging of dementia,'' \emph{The British journal of
  psychiatry}, vol. 140, no.~6, pp. 566--572, 1982.

\bibitem{rosen1984new}
W.~G. Rosen, R.~C. Mohs, and K.~L. Davis, ``A new rating scale for alzheimer's
  disease.'' \emph{The American journal of psychiatry}, 1984.

\bibitem{weiner2017recent}
M.~W. Weiner, D.~P. Veitch, P.~S. Aisen, L.~A. Beckett, N.~J. Cairns, R.~C.
  Green, D.~Harvey, C.~R. Jack~Jr, W.~Jagust, J.~C. Morris \emph{et~al.},
  ``Recent publications from the alzheimer's disease neuroimaging initiative:
  Reviewing progress toward improved ad clinical trials,'' \emph{Alzheimer's \&
  Dementia}, vol.~13, no.~4, pp. e1--e85, 2017.

\bibitem{rasmussen2006cki}
C.~Rasmussen, ``Cki williams gaussian processes for machine learning,'' 2006.

\bibitem{candela2003propagation}
J.~Q. Candela, A.~Girard, J.~Larsen, and C.~E. Rasmussen, ``Propagation of
  uncertainty in bayesian kernel models-application to multiple-step ahead
  forecasting,'' in \emph{2003 IEEE International Conference on Acoustics,
  Speech, and Signal Processing, 2003. Proceedings.(ICASSP'03).}, vol.~2.\hskip
  1em plus 0.5em minus 0.4em\relax IEEE, 2003, pp. II--701.

\bibitem{eleftheriadis2017gaussian}
S.~Eleftheriadis, O.~Rudovic, M.~P. Deisenroth, and M.~Pantic, ``Gaussian
  process domain experts for modeling of facial affect,'' \emph{IEEE
  transactions on image processing}, vol.~26, no.~10, pp. 4697--4711, 2017.

\bibitem{liu2015bayesian}
B.~Liu and N.~Vasconcelos, ``Bayesian model adaptation for crowd counts,'' in
  \emph{Proceedings of the IEEE International Conference on Computer Vision},
  2015, pp. 4175--4183.

\bibitem{cox1972regression}
D.~R. Cox, ``Regression models and life-tables,'' \emph{Journal of the Royal
  Statistical Society: Series B (Methodological)}, vol.~34, no.~2, pp.
  187--202, 1972.

\bibitem{utsumil2018personalized}
Y.~Utsumil, O.~O. Rudovicl, K.~Petersonl, R.~Guerrero, and R.~W. Picardl,
  ``Personalized gaussian processes for forecasting of alzheimer’s disease
  assessment scale-cognition sub-scale (adas-cog13),'' in \emph{2018 40th
  Annual International Conference of the IEEE Engineering in Medicine and
  Biology Society (EMBC)}.\hskip 1em plus 0.5em minus 0.4em\relax IEEE, 2018,
  pp. 4007--4011.

\bibitem{schmidt2016learning}
A.~Schmidt-Richberg, C.~Ledig, R.~Guerrero, H.~Molina-Abril, A.~Frangi,
  D.~Rueckert, A.~D.~N. Initiative \emph{et~al.}, ``Learning biomarker models
  for progression estimation of alzheimer’s disease,'' \emph{PloS one},
  vol.~11, no.~4, p. e0153040, 2016.

\bibitem{guerrero2016instantiated}
R.~Guerrero, A.~Schmidt-Richberg, C.~Ledig, T.~Tong, R.~Wolz, D.~Rueckert,
  A.~D. N.~I. (ADNI \emph{et~al.}, ``Instantiated mixed effects modeling of
  alzheimer's disease markers,'' \emph{NeuroImage}, vol. 142, pp. 113--125,
  2016.

\bibitem{williams1996gaussian}
C.~K. Williams and C.~E. Rasmussen, ``Gaussian processes for regression,'' in
  \emph{Advances in neural information processing systems}, 1996, pp. 514--520.

\bibitem{neal1993bayesian}
R.~M. Neal, ``Bayesian learning via stochastic dynamics,'' in \emph{Advances in
  neural information processing systems}, 1993, pp. 475--482.

\bibitem{mml}
C.~S.~O. Marc P~Desienroth, A. Aldo~Faisal, \emph{Mathematics for Machine
  Learning}.\hskip 1em plus 0.5em minus 0.4em\relax Cambridge University Press,
  2020.

\bibitem{d1995multidimensional}
G.~D'Agostini, ``A multidimensional unfolding method based on bayes' theorem,''
  \emph{Nuclear Instruments and Methods in Physics Research Section A:
  Accelerators, Spectrometers, Detectors and Associated Equipment}, vol. 362,
  no. 2-3, pp. 487--498, 1995.

\bibitem{welch1995introduction}
B.~G. Welch, Greg, ``An introduction to the kalman filter,'' 1995.

\bibitem{willmott2005advantages}
C.~J. Willmott and K.~Matsuura, ``Advantages of the mean absolute error (mae)
  over the root mean square error (rmse) in assessing average model
  performance,'' \emph{Climate research}, vol.~30, no.~1, pp. 79--82, 2005.

\bibitem{kanungo2002efficient}
T.~Kanungo, D.~M. Mount, N.~S. Netanyahu, C.~D. Piatko, R.~Silverman, and A.~Y.
  Wu, ``An efficient k-means clustering algorithm: Analysis and
  implementation,'' \emph{IEEE Transactions on Pattern Analysis \& Machine
  Intelligence}, no.~7, pp. 881--892, 2002.

\bibitem{ng2002spectral}
A.~Y. Ng, M.~I. Jordan, and Y.~Weiss, ``On spectral clustering: Analysis and an
  algorithm,'' in \emph{Advances in neural information processing systems},
  2002, pp. 849--856.

\bibitem{merris1994laplacian}
R.~Merris, ``Laplacian matrices of graphs: a survey,'' \emph{Linear algebra and
  its applications}, vol. 197, pp. 143--176, 1994.

\bibitem{bello2019deep}
G.~A. Bello, T.~J. Dawes, J.~Duan, C.~Biffi, A.~de~Marvao, L.~S. Howard,
  J.~S.~R. Gibbs, M.~R. Wilkins, S.~A. Cook, D.~Rueckert \emph{et~al.},
  ``Deep-learning cardiac motion analysis for human survival prediction,''
  \emph{Nature machine intelligence}, vol.~1, no.~2, p.~95, 2019.

\bibitem{faraggi1995neural}
D.~Faraggi and R.~Simon, ``A neural network model for survival data,''
  \emph{Statistics in medicine}, vol.~14, no.~1, pp. 73--82, 1995.

\bibitem{cheadle2003analysis}
C.~Cheadle, M.~P. Vawter, W.~J. Freed, and K.~G. Becker, ``Analysis of
  microarray data using z score transformation,'' \emph{The Journal of
  molecular diagnostics}, vol.~5, no.~2, pp. 73--81, 2003.

\bibitem{suykens1999least}
J.~A. Suykens and J.~Vandewalle, ``Least squares support vector machine
  classifiers,'' \emph{Neural processing letters}, vol.~9, no.~3, pp. 293--300,
  1999.

\bibitem{safavian1991survey}
S.~R. Safavian and D.~Landgrebe, ``A survey of decision tree classifier
  methodology,'' \emph{IEEE transactions on systems, man, and cybernetics},
  vol.~21, no.~3, pp. 660--674, 1991.

\bibitem{kleinbaum2002logistic}
D.~G. Kleinbaum, K.~Dietz, M.~Gail, M.~Klein, and M.~Klein, \emph{Logistic
  regression}.\hskip 1em plus 0.5em minus 0.4em\relax Springer, 2002.

\end{thebibliography}
\end{document}